\documentclass{article}

\usepackage{PRIMEarxiv}

\usepackage[utf8]{inputenc} 
\usepackage[T1]{fontenc}    
\usepackage{hyperref}       
\usepackage{url}            
\usepackage{booktabs}       
\usepackage{amsfonts}       
\usepackage{nicefrac}       
\usepackage{microtype}      
\usepackage{wrapfig}
\usepackage{lipsum}
\usepackage{paracol}
\usepackage{caption}
\usepackage{adjustbox}
\usepackage{float}
\usepackage{placeins}
\usepackage{fancyhdr}       
\usepackage{graphicx}       
\graphicspath{{media/}}     

\usepackage[sort]{natbib}
\setcitestyle{super, numbers,open={[},close={]}}
\usepackage{xurl}
\title{Automatic Detection of Gen-AI Texts: A Comparative Framework of Neural Models}

\author{
  Cristian Buttaro\\
  Sapienza University in Rome \\
  Latina\\
  \texttt{buttaro.2071786@studenti.uniroma1.it} \\
  \And
  Irene Amerini \\
  Sapienza University in Rome \\
  Latina \\
  \texttt{irene.amerini@uniroma1.it} \\
}

\begin{document}
\twocolumn[
\begin{@twocolumnfalse}
\maketitle

\begin{abstract}
The rapid proliferation of Large Language Models (LLMs) has significantly increased the difficulty of distinguishing between human-written and AI-generated texts, raising critical issues across academic, editorial, and social domains. This paper investigates the problem of \emph{AI-generated text detection} through the design, implementation, and comparative evaluation of multiple machine learning–based detectors. Four neural architectures are developed and analyzed: a Multilayer Perceptron (MLP), a one-dimensional Convolutional Neural Network (CNN 1D), a MobileNet-based CNN, and a Transformer model. The proposed models are benchmarked against widely used online detectors, including ZeroGPT \cite{zerogpt}, GPTZero \cite{gptzero}, QuillBot \cite{quillbot}, Originality.AI \cite{originality}, Sapling \cite{sapling}, IsGen \cite{isgen}, Rephrase \cite{rephrase}, and Writer \cite{writer}. Experiments are conducted on the \emph{COLING Multilingual Dataset} \cite{genaidetect2025dataset}, considering both English and Italian configurations, as well as on an original thematic dataset focused on Art and Mental Health. Results show that supervised detectors achieve more stable and robust performance than commercial tools across different languages and domains, highlighting key strengths and limitations of current detection strategies.
\end{abstract}
\vspace{10pt}
\end{@twocolumnfalse}]
\section{Introduction}
In recent years, generative artificial intelligence has profoundly transformed the production and circulation of textual content. Large Language Models (LLMs) have achieved a level of fluency and coherence that makes it increasingly difficult to distinguish artificially generated texts from those written by humans \cite{goodfellow2016dl,prince2023udl,brown2020language,openai2023gpt4}.\\The growing accessibility of these tools has led to an exponential increase in AI-generated content across educational, journalistic, administrative and legal domains, raising significant concerns regarding reliability, transparency, and accountability.\\
In response to this scenario, a dedicated line of research has emerged focusing on the \emph{detection of AI-generated texts}, positioned at the intersection of computational linguistics, machine learning, and multimedia forensics.\\
Nevertheless, despite the variety of proposed approaches, reliably distinguishing between human-written and AI-generated texts remains an open challenge. \\The main detection strategies include stylistic and linguistic analysis \cite{gehrmann2019gltr}, methods based on token-level probability and log-likelihood curvature, statistical watermarking techniques \cite{kirchenbauer2023watermarking}, and supervised classifiers trained on balanced \emph{Human-GenAI} datasets \cite{zellers2019defending,liu2024comparative}. Each approach exhibits structural limitations, particularly in terms of generalization, cross-model robustness, and susceptibility to false positives and false negatives.\\
These limitations are not merely technical, but also give rise to significant social, ethical, and legal consequences.\\
Recent studies have shown that detection errors may lead to false accusations, discrimination, and a loss of trust in educational, media, and judicial institutions \cite{weidinger2022taxonomy,hayawi2023imitation}. \\Several episodes reported in the Italian context, spanning academic, media, and legal settings, illustrate how the uncritical adoption of detection tools can result in arbitrary and potentially unfair decisions \cite{repubblica2025studentessa,latinaoggi2025,ilcaffe2025}.\\
In light of these challenges, text detection cannot be treated as a simple automated classification problem, but instead requires a scientifically rigorous and socially responsible approach, aligned with the ongoing European regulatory debate (AI Act, GDPR). Within this context, the present work aims to analyze existing AI-generated text detection methodologies, assess the reliability of widely used commercial tools, and propose a supervised experimental detector evaluated under realistic multilingual and domain-specific conditions \cite{genaidetect2025dataset,liu2024comparative}. The ultimate goal is to provide empirical insights that support a more reliable and responsible distinction between human-written and AI-generated texts.\\
In line with open science principles and to facilitate reproducibility, all datasets, experimental materials, and implementation details are publicly available.\footnote{\url{https://github.com/cristian03git/DETECTION_GENAI.git}}

\section{Related Works}
The detection of AI-generated text is a relatively recent yet rapidly evolving research area, characterized by a growing body of academic contributions and the parallel emergence of commercial detection tools. \\Existing works have explored this problem along multiple methodological directions, including stylistic and linguistic analysis, probabilistic approaches, supervised classifiers, statistical watermarking, and cognitive perspectives.\\
Early supervised models combined and entropic signals to discriminate between human-written and synthetic texts, while subsequent studies demonstrated that token predictability and distributional irregularities can serve as effective indicators of artificial generation \cite{gehrmann2019gltr,zellers2019grover}. \\With the advent of Transformer-based language models, several works showed that latent contextual representations capture syntactic and semantic cues useful for \emph{Human-GenAI} discrimination \cite{ippolito2020automatic}. Parallel research also addressed the ethical and societal implications of automated text generation and detection \cite{solaiman2019release}. More recent approaches have introduced increasingly sophisticated detection strategies.\\ 
DetectGPT exploits curvature-based properties of token-level log probabilities \cite{mitchell2023detectgpt}, while watermarking techniques propose embedding imperceptible statistical signatures into generated text \cite{kirchenbauer2023watermarking}. Comparative studies consistently report a growing difficulty in detection as language models improve, as well as substantial variability and limited reliability among commercial detectors \cite{hayawi2023imitation,elkhatat2023detection}. \\Since 2024, research has increasingly focused on application-specific and robustness-oriented evaluations. Studies in educational and medical contexts have highlighted the risks associated with false positives and the social consequences of unreliable detection \cite{chowdhury2024genaidetect,durak2025comparison,doru2025medical}. \\Other works have investigated cross-model generalization, hybrid \emph{Human-GenAI} texts, and multilingual or domain-shift scenarios, revealing persistent limitations in robustness and generalization \cite{guo2024detective,zeng2024hybrid,liu2024comparative}. Alongside academic research, a broad ecosystem of online detectors has emerged, including ZeroGPT, GPTZero, QuillBot, Writer, Sapling, Originality.AI, IsGen, and Rephrase. \\Despite their widespread adoption, these tools often lack methodological transparency and exhibit high error rates, reinforcing a persistent dichotomy between academic approaches and opaque real-world systems \cite{elkhatat2023detection,liu2024comparative}.\\
A large-scale comparative evaluation of detection systems is presented in \cite{genaidetect2025dataset}, where numerous approaches, primarily based on fine-tuned large language models and ensemble strategies, are assessed under a fixed shared training and evaluation protocol. In contrast, the present work at a controlled and architecture-centered analysis of detection stability across languages and domains.\\
Despite the rapid growth of AI-generated text detection research, important gaps remain. 
Many studies focus on single-language (often English-only) and balanced benchmarks, limiting insight into multilingual behavior and domain variability. 
Moreover, academic models and commercial detectors are typically evaluated separately, resulting in a limited understanding of their reliability under consistent conditions.\\
This work addresses these limitations through a unified comparative framework. 
We design and evaluate supervised neural detectors based on heterogeneous architectures, like feed-forward, convolutional, and Transformer-based, across four controlled scenarios defined by language (English and Italian) and dataset typology (general-purpose and thematic). 
Unlike prior studies that emphasize performance, we explicitly investigate cross-lingual stability and domain sensitivity. 
The proposed models are further benchmarked against widely used commercial detectors under the same protocol, providing an assessment of robustness and reliability across heterogeneous evaluation settings.

\section{Methodology}
\label{sec:prop-method}
This work proposes a modular and comparable framework for binary \emph{Human vs.\ GenAI} text classification, in which all detectors share the same end-to-end pipeline and differ only in the neural \emph{feature extraction} module.
\begin{figure}[t]
    \centering
    \includegraphics[width=0.80\linewidth]{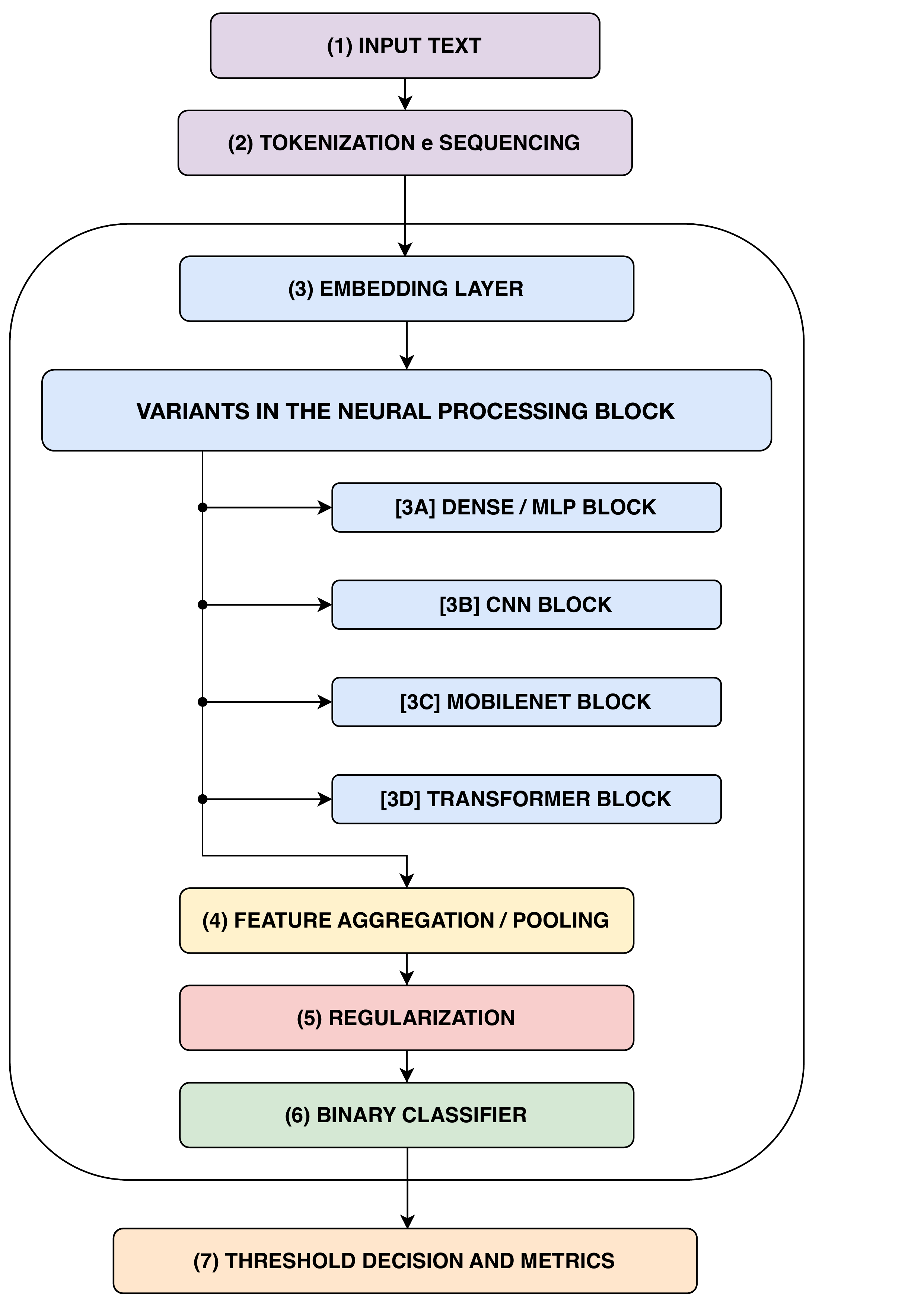}
    \caption{General pipeline of the text detection system.}
    \label{fig:INGpipeline_generale}
\end{figure}
\noindent
Figure~\ref{fig:INGpipeline_generale} provides an overview of the proposed end-to-end \emph{Human vs.\ GenAI} detection pipeline. Given a raw input text $x_i$, the system produces a fixed-length numerical representation through the following stages:
\begin{enumerate}
    \item tokenization and sequencing, converting text into token IDs and normalizing
    sequences to a maximum length $L$ via padding or truncation;
    \item an embedding layer, yielding a dense matrix representation
    $E \in \mathbb{R}^{L \times d}$;
    \item a neural feature extractor, generating contextual or convolutional feature
    maps $H \in \mathbb{R}^{L \times k}$;
    \item global feature aggregation through pooling, producing a fixed-size vector $h \in \mathbb{R}^{k}$;
    \item regularization with dropout to mitigate overfitting;
    \item a binary classification head, outputting a probability score $\hat{y} \in [0,1]$, followed by a threshold-based decision $\tau$.
\end{enumerate}
This final component is empirically calibrated on validation data to balance sensitivity and specificity, reducing false positives on highly polished human texts. The core methodological comparison focuses on four model families:

\begin{itemize}
    \item MLP (Dense Networks). Used as a lightweight baseline, the MLP operates on an aggregated representation of the sequence obtained via \emph{masked pooling} over token embeddings. The pooled vectors are concatenated and passed through a compact MLP head with ReLU and dropout, providing a stable reference model without explicit sequence modeling \cite{hornik1989multilayer,rumelhart1986learning,vaswani2017attention}.
    \item CNN 1D. Convolutional detectors apply 1D filters directly over the embedding sequence to capture \emph{local patterns} corresponding to short contiguous groups of tokens (i.e., patterns analogous to traditional n-grams in statistical language modeling). A single convolutional layer generates feature maps that are aggregated using Global Max Pooling, emphasizing salient local cues commonly associated with synthetic text, followed by dropout and a sigmoid-based classifier \cite{lecun1998gradient,kim2014convolutional}.
    \item MobileNet-based 1D CNN. To improve parameter efficiency, this detector employs 1D depthwise-separable convolutions, following the computational design principle of MobileNet \cite{howard2017mobilenets}. Unlike the original 2D vision model, convolutions operate over token embeddings, making the architecture suitable for sequential text data. The model is tailored for long English sequences and uses a larger embedding dimension to mitigate the representational compression introduced by separable convolutions. Feature aggregation combines global average and max pooling, capturing both distributional trends and peak activations.
    \item Transformer. The transformer-based detector models \emph{long-range contextual dependencies} via multi-head self-attention \cite{vaswani2017attention}. Token embeddings are augmented with positional information, processed by stacked encoder blocks (attention and feed-forward layers with LayerNorm and dropout), and summarized using a combination of pooling strategies. The resulting global representation is passed to a fully connected classification head and thresholded to produce the final decision.\\
\end{itemize}
Beyond architectural differences, the comparison also considers hyperparameter configurations.\\ 
For the MLP-based detectors, embedding and hidden dimensions are fixed to 128 across datasets to ensure comparability. Regularization and calibration are supported through dropout (0.20–0.30), label smoothing (up to 0.05), weight decay ($10^{-4}$ – $2\times10^{-4}$), and validation-based threshold tuning ($\tau \in [0.35,0.40]$).\\
The CNN 1D models adapt embedding size (128–300), number of filters (128–400), and kernel configurations according to dataset scale. Larger capacity and batch sizes are adopted for \emph{dtEN} dataset, while more compact settings are used for \emph{dtITA} dataset. Decision thresholds are either validation-optimized ($\tau \approx 0.35$–$0.42$) or derived via argmax.
The CNN Mobilenet employs embedding dimension 256, maximum sequence length 1024, batch size 192, learning rate $2\times10^{-4}$, weight decay (0.01), label smoothing (0.05), and 8 training epochs with validation-based threshold calibration ($\tau=0.36$).\\
The Transformer-based detector consists of stacked encoder layers (with 8 attention heads and feed-forward dimension 1024 per block), embedding dimension 256, maximum sequence length 1024, and batch size 192. Training is performed for 8 epochs with reduced learning rate ($2\times10^{-4}$), weight decay (0.01), dropout (0.10), and label smoothing (0.05), using validation monitoring for threshold calibration ($\tau=0.36$) and convergence control.\\
In addition to the proposed models, the study includes a methodological comparison with widely used online detectors, such as ZeroGPT \cite{zerogpt}, GPTZero \cite{gptzero}, QuillBot \cite{quillbot}, Originality.AI \cite{originality}, Sapling \cite{sapling}, IsGen \cite{isgen}, Rephrase \cite{rephrase}, and Writer \cite{writer}. Although their internal architectures are not publicly disclosed, these tools typically rely on proprietary combinations of perplexity-based scoring, stylometric features, burstiness analysis, and large-scale supervised classifiers trained to distinguish human and LLM-generated text. They are evaluated with respect to detection behavior, and potential failure modes (e.g., false positives), positioning the proposed framework against practical solutions currently adopted in real-world settings.

\section{Overview Dataset}
Two main data sources were considered a selected portions of the \emph{COLING Multilingual Dataset} and a set of \emph{original thematic datasets} specifically designed within this work. The first dataset source is derived from the \emph{GenAI Content Detection Task 1} \cite{genaidetect2025dataset} organized at COLING 2025. This benchmark was selected due to its multilingual coverage, and diversity of generative sources. \\An additional motivation for this choice is to systematically assess how widely used online detectors, which are predominantly optimized for English, perform when applied to non-English languages. The dataset~\cite{genaidetect2025dataset} is publicly available via Hugging Face \footnote{\url{https://huggingface.co/datasets/Jinyan1/COLING_2025_MGT_multingual}}.\\
Each record includes metadata such as source, language, generative model, binary label (\emph{Human vs.\ GenAI}), and the text itself. From this resource, two subsets were extracted. The \emph{dtEN} subset contains English texts with both \emph{Human} and \emph{GenAI} samples and serves as the primary large-scale benchmark for binary detection. In contrast, the \emph{dtITA} subset consists of Italian texts which, in the original release, include only \emph{GenAI} samples; this configuration enables the analysis of single-class settings and the evaluation of dataset balancing strategies, as well as a focused investigation of detector behavior in a language other than English.\\
In addition to public benchmarks, a set of thematic Italian datasets, called \emph{ART\&MH}, was constructed to assess detector robustness in semantically specific and stylistically complex domains. Two thematic domains were selected: mental health and artwork descriptions. \\These domains were chosen to test detection performance on narrative texts (mental health) and on descriptive and interpretative content (art). For each topic, both \emph{GenAI} texts, produced using Gemini~2.5~Flash, Claude~Sonnet~4, and GPT-4.5, and human-written texts were created. Each dataset is split into training, validation, and test sets following standard supervised learning practice. \\Unlike the COLING-derived datasets, the thematic datasets adopt a minimal structure consisting solely of the text and its binary label. Representative examples are reported in Tables~\ref{tab:arte} and~\ref{tab:sm}, illustrating stylistic differences between \emph{Human} and \emph{GenAI} samples in the Art and Mental Health domains, respectively. \\All datasets undergo the same preprocessing, tokenization, and sequencing pipeline described in Section~\ref{sec:prop-method}, and are used to train and evaluate the detectors proposed in this work.
\begin{table}
    \caption{Example records from the \emph{Art} topic.
    Label~0 denotes human-written text, while Label~1 denotes GenAI text.}
    \label{tab:arte}
    \centering
    \begin{tabular}{p{5.5cm} c}
        \toprule
        \textbf{Text} & \textbf{Label} \\
        \midrule
        Gli Amanti (Magritte)\\
        Questo quadro mi ha sempre fatto un effetto strano. Due persone che si baciano ma hanno la testa coperta da dei teli... è romantico e inquietante allo stesso tempo. [...]
        & 0 \\
        \midrule
        Il Bacio (Hayez)\\
        L'opera di Francesco Hayez incarna l'ideale romantico dell'amore patriottico, dove il gesto intimo tra i due amanti assume significati politici legati al Risorgimento italiano. [...]
        & 1 \\
        \bottomrule
    \end{tabular}
\end{table}
\begin{table}
    \caption{Example records from the \emph{Mental Health} topic.
    Label~0 denotes human-written text, while Label~1 denotes GenAI text.}
    \label{tab:sm}
    \centering
    \begin{tabular}{p{5.5cm} c}
        \toprule
        \textbf{Text} & \textbf{Label} \\
        \midrule
        Era un martedì mattina qualunque. Stavo preparando la colazione e all'improvviso ho iniziato a piangere senza motivo. Non riuscivo a smettere. [...]
        & 0 \\
        \midrule
        Secondo gli studi, la prevenzione del suicidio funziona meglio quando si usano controlli regolari e aiuti immediati durante le crisi. [...]
        & 1 \\
        \bottomrule
    \end{tabular}
\end{table}
\noindent
All experiments were conducted on test sets composed of 60 samples per dataset. Performance is reported in terms of overall accuracy and class-wise detection rates for \emph{Human} and \emph{GenAI} texts.\\
The choice of 60 samples per setting was intentional and aimed at ensuring controlled and comparable evaluations across datasets and detectors. Each subset was balanced and manually verified, privileging data quality and annotation reliability over scale. \\Moreover, results are observed across different datasets and experimental scenarios, so the consistency of trends across settings mitigates the limitations typically associated with smaller test partitions.
\subsection{Results on dtEN dataset}
The \emph{dtEN} dataset represents a balanced English-language scenario with moderate stylistic variability.\\ Table~\ref{tab:results_dten} summarizes the results obtained by the implemented detectors and by online tools.
\begin{table}
    \caption{Results on the \emph{dtEN} dataset.}
    \label{tab:results_dten}
    \centering
    \small
    \begin{tabular}{l c c c}
        \toprule
        \textbf{Detector} & \textbf{Accuracy(\%)}& \textbf{Human(\%)}& \textbf{GenAI(\%)}\\
        \midrule
        MLP &85.0&97.1&68.0\\
        CNN 1D  &70.0&0.0&100.0\\
        MobileNet CNN & \textbf{91.67}&83.33&95.24\\
        Transformer  &88.3&97.3&73.9\\
        \midrule
        GPTZero  & \underline{90.0}&100.0&81.2\\
        Sapling  &83.3&100.0&68.8\\
        Originality  &80.0&100.0&62.5\\
        IsGen   &76.7&100.0&56.2\\
        ZeroGPT  &68.3&92.9&46.9\\
        QuillBot  &65.0&100.0&34.4\\
        Rephrase  &63.3&96.4&34.4\\
        Writer  &53.3&0.0&100.0\\
        \bottomrule
    \end{tabular}
\end{table}
\noindent
No detector achieves perfect separation between \emph{Human} and \emph{GenAI} texts, confirming the intrinsic ambiguity of the task. Among the proposed models, the MobileNet CNN achieves the best overall trade-off, combining high sensitivity to \emph{GenAI} texts with a reasonable preservation of human samples.\\
The MLP and Transformer models instead exhibit a more conservative behavior, characterized by very high accuracy on human-written texts (97.1\% and 97.3\%, respectively). This suggests a bias toward minimizing false positives at the expense of missing a fraction of AI-generated content.
Conversely, the CNN~1D collapses toward the \emph{GenAI} class, yielding perfect \emph{GenAI} detection but completely failing to recognize human texts, which highlights the limitations of relying exclusively on local convolutional features in this setting. 
Online detectors often show high accuracy on human texts but substantially lower sensitivity to \emph{GenAI} content, indicating a systematic tendency to prioritize false-positive avoidance.
Since these commercial detectors were not specifically trained on the \emph{dtEN} subset, their results provide an indication of cross-dataset generalization capability.
\subsection{Results on the dtITA dataset}
The \emph{dtITA} dataset contains only  Italian \emph{GenAI} texts and represents a single-class evaluation scenario. In this setting, accuracy reflects the proportion of correctly identified \emph{GenAI} samples, while any prediction of the \emph{Human} class corresponds to a misclassification. Results are reported in Table~\ref{tab:results_dtita}.
\begin{table}[]
    \caption{Results on the \emph{dtITA} dataset.}
    \label{tab:results_dtita}
    \centering
    \small
    \begin{tabular}{l c c c}
        \toprule
        \textbf{Detector} & \textbf{Accuracy (\%)} & \textbf{Human (\%)} & \textbf{GenAI (\%)} \\
        \midrule
        MLP        & \textbf{100.0} & 0.0  & 100.0 \\
        CNN 1D     & \textbf{100.0} & 0.0  & 100.0 \\
        \midrule
        Writer      & \textbf{100.0} & 0.0  & 100.0 \\
        Rephrase    & \underline{80.0}  & 20.0 & 80.0 \\
        QuillBot    & 76.7  & 23.3 & 76.7 \\
        Sapling     & 75.0  & 25.0 & 75.0 \\
        GPTZero     & 61.7  & 38.3 & 61.7 \\
        ZeroGPT     & 56.7  & 43.3 & 56.7 \\
        IsGen       & 56.7  & 43.3 & 56.7 \\
        Originality & 53.3  & 46.7 & 53.3 \\
        \bottomrule
    \end{tabular}
\end{table}
\noindent
The MobileNet-style CNN and the Transformer-based detector are not evaluated in this scenario, as the \emph{dtITA} dataset contains a limited number of samples and only \emph{GenAI} instances. Such a small and single-class setting would not allow effective training or meaningful evaluation of high-capacity architectures. For this reason, the analysis focuses on lightweight supervised detectors and commercial tools, whose behavior under distributional shift can be more clearly interpreted.
The implemented detectors correctly classify all \emph{GenAI} samples, exhibiting stable decision behavior even in the absence of \emph{Human} examples. This outcome indicates that, in a single-class setting, the proposed models maintain consistent classification behavior on \emph{GenAI} samples. \\In contrast, several online detectors show a marked degradation in performance, misclassifying a substantial portion of \emph{GenAI} texts as \emph{Human}, highlighting limited robustness under distributional shift.

\subsection{Cross-Domain Test on dtITA}
To further assess robustness, \emph{dtITA} was used as a single-class test set for models trained on different datasets.\\ 
In Table~\ref{tab:monoclass_dtita_results}, each model is reported together with the corresponding training dataset to explicitly highlight the effect of training data on single-class generalization performance.
\begin{table}[t]
    \caption{Cross-domain single-class evaluation on the \emph{dtITA} dataset. Parentheses indicate the training dataset of each detector.}
    \label{tab:monoclass_dtita_results}
    \centering
    \footnotesize
    \setlength{\tabcolsep}{3.5pt}
    \renewcommand{\arraystretch}{0.95}
    \begin{tabular}{lccc}
    \toprule
    \textbf{Detector} & \textbf{Accuracy(\%)} & \textbf{Human(\%)} & \textbf{GenAI(\%)} \\
    \midrule
    CNN 1D (\emph{ART\&MH}) & \textbf{92.35} & 0.0 & 92.35 \\
    MLP (\emph{ART\&MH}) & 90.07 & 0.0 & 90.07 \\
    MLP (\emph{dtEN}) & \underline{91.32} & 0.0 & 91.32 \\
    Transformer (\emph{dtEN}) & 88.66 & 0.0 & 88.66 \\
    MobileNet (\emph{dtEN}) & 85.35 & 0.0 & 85.35 \\
    \bottomrule
    \end{tabular}
\end{table}
Models trained on the heterogeneous \emph{ART\&MH} dataset, which is also composed of Italian texts, exhibit stronger cross-domain robustness, achieving higher accuracy in identifying \emph{GenAI} content under language shift.\\ 
This suggests that both exposure to stylistically diverse data and linguistic alignment with the target language contribute to improved generalization in single-class evaluation settings.\\
Conversely, architectures optimized on the English \emph{dtEN} dataset show a more pronounced performance degradation when evaluated on Italian texts, particularly for deeper models. This behavior highlights sensitivity to language-specific statistical patterns and reduced robustness under cross-lingual distributional shift.
\subsection{Results on thematic Dataset ART\&MH}
The \emph{ART\&MH} dataset includes highly variable human texts related to art and mental health, representing a challenging detection scenario.\\
Results are summarized in Table~\ref{tab:results_artmh}.

\begin{table}[t]
    \caption{Results on the \emph{ART\&MH} thematic dataset.}
    \label{tab:results_artmh}
    \centering
    \small
    \begin{tabular}{l c c c}
        \toprule
        \textbf{Detector} & \textbf{Accuracy(\%)} & \textbf{Human(\%)}& \textbf{GenAI(\%)}\\
        \midrule
        CNN 1D  & \underline{98.3}&96.8&100.0\\
        MLP   & \underline{98.3} &100.0&96.8\\
        \midrule
        ZeroGPT  & \textbf{100.0}&100.0&100.0\\
        GPTZero  & \textbf{100.0}&100.0&100.0\\
        QuillBot  & \textbf{100.0}&100.0&100.0\\
        Originality & \textbf{100.0}&100.0&100.0\\
        Sapling  &98.3&100.0&96.7\\
        IsGen  &93.3&100.0&86.7\\
        Rephrase &76.7&100.0&53.3\\
        Writer &50.0&100.0&0.0\\
        \bottomrule
    \end{tabular}
\end{table}
\noindent
The proposed detectors achieve high performance while maintaining balanced behavior across \emph{Human} and \emph{GenAI} classes. 
The MLP prioritizes the preservation of human-written texts by minimizing false positives, whereas the CNN~1D emphasizes the identification of \emph{GenAI} content, at the cost of reduced discrimination in certain scenarios.\\
The Writer detector collapses all predictions toward the \emph{Human} class, completely failing to identify \emph{GenAI} texts. Other commercial tools achieve high accuracy on this dataset without exhibiting the same behavior.\\
Model behavior depends on the decision threshold $\tau$, probability calibration, and regularization, in addition to the underlying architecture:
\begin{itemize}
    \item on \emph{dtEN}, a clear trade-off emerges between minimizing false positives on human-written texts and maintaining sensitivity to \emph{GenAI} content, distinguishing more conservative models from more balanced detection approaches;
    \item on the \emph{monoclass dtITA} setting, the implemented detectors exhibit stable behavior, whereas several online tools reveal a bias toward the \emph{Human} class;
    \item the \emph{cross-domain test} on \emph{dtITA} highlights the impact of \emph{dataset shift}, with better transfer observed for models trained on more heterogeneous domains (\emph{ART\&MH}) compared to those optimized on a single domain (\emph{dtEN});
    \item finally, on \emph{ART\&MH}, the proposed models maintain high performance while making different types of errors, with some favoring human-text preservation and others emphasizing \emph{GenAI} detection. In contrast, some online tools achieve seemingly perfect results that are not always interpretable due to the lack of transparency regarding thresholds and calibration.
\end{itemize}
\section{Conclusions}
This work addressed the problem of \emph{Human-GenAI text detection}, providing a systematic analysis of different supervised neural architectures and comparing them with widely used online detection tools.\\
The study investigated the effectiveness of MLP, CNN 1D, CNN MobileNet, and Transformer-based detectors across multiple datasets characterized by different languages, and stylistic variability. Experimental results show that no universally optimal detector exists.
Instead, model behavior depends critically on architectural choices as well as on decision thresholds, probability calibration, and regularization strategies.
Balanced datasets such as \emph{dtEN} highlight trade-offs between conservative models that preserve human texts and models that are more sensitive to \emph{GenAI} content, while thematic data (\emph{ART\&MH}) expose the difficulty of distinguishing highly expressive human writing from synthetic text.\\
The monoclass and cross-domain experiments on \emph{dtITA} further demonstrate that robustness under distributional shift cannot be reliably assessed using standard balanced evaluations alone, as models that perform well on in-domain, balanced test sets may exhibit significant degradation when applied to data from different languages or domains.\\
A meaningful evaluation therefore requires controlled experiments conducted under heterogeneous conditions and stress-test scenarios that explicitly probe robustness beyond standard in-domain settings.\\
Future work will focus on extending multilingual coverage and systematically analyzing language shift across additional languages and subdomains. \\Further research will explore ensemble and hybrid detection strategies that combine heterogeneous architectural paradigms to improve robustness and generalization under complex distributional conditions.\\
Another promising direction concerns threshold calibration and adaptive decision mechanisms, aimed at reducing false positives in sensitive application domains. Investigating uncertainty estimation and confidence-aware prediction strategies may further enhance the practical reliability of detection systems. 
From an application perspective, future developments may include the integration of the proposed models into real-world software for educational, professional, and domain-specific use cases. \\In particular, thematic datasets such as \emph{ART\&MH} suggest potential intersections with language-based analysis in mental health contexts, where robust and transparent detection mechanisms could support broader AI-assisted assessment frameworks.\\
This study contributes to a clearer understanding of both the potential and the current limitations of \emph{GenAI} detection systems, emphasizing the importance of transparency, robustness, and domain-aware evaluation.

\bibliographystyle{unsrt}  
\bibliography{references}

\end{document}